\documentclass[runningheads]{llncs}
\usepackage[T1]{fontenc}
\usepackage{graphicx}
\usepackage[misc]{ifsym}
\begin{document}
\title{Explaining Full-disk Deep Learning Model for Solar Flare Prediction using Attribution Methods}
\titlerunning{Explaining Full-disk Solar Flare Prediction Model}
%
%
\author{Chetraj Pandey\Letter \orcidID{0000-0002-4699-4050} \and Rafal A. Angryk\orcidID{0000-0001-9598-8207} \and
Berkay Aydin\orcidID{0000-0002-9799-9265}}
\authorrunning{C. Pandey et al.}
%
\institute{Georgia State University, Atlanta, GA, 30303, USA\\
\email{\{cpandey1, rangryk, baydin2\} @gsu.edu}}
\toctitle{Explaining Full-disk Deep Learning Models for Solar Flare Prediction using Attribution Methods}
\tocauthor{Chetraj~Pandey, Rafal A.~Angryk, Berkay~Aydin}

\maketitle              
\begin{abstract}
Solar flares are transient space weather events that pose a significant threat to space and ground-based technological systems, making their precise and reliable prediction crucial for mitigating potential impacts. This paper contributes to the growing body of research on deep learning methods for solar flare prediction, primarily focusing on highly overlooked near-limb flares and utilizing the attribution methods to provide a post hoc qualitative explanation of the model's predictions. We present a solar flare prediction model, which is trained using hourly full-disk line-of-sight magnetogram images and employs a binary prediction mode to forecast $\geq$M-class flares that may occur within the following 24-hour period. To address the class imbalance, we employ a fusion of data augmentation and class weighting techniques; and evaluate the overall performance of our model using the true skill statistic (TSS) and Heidke skill score (HSS). Moreover, we applied three attribution methods, namely Guided Gradient-weighted Class Activation Mapping, Integrated Gradients, and Deep Shapley Additive Explanations, to interpret and cross-validate our model's predictions with the explanations. Our analysis revealed that full-disk prediction of solar flares aligns with characteristics related to active regions (ARs). In particular, the key findings of this study are: (1) our deep learning models achieved an average TSS$\sim$0.51 and HSS$\sim$0.35, and the results further demonstrate a competent capability to predict near-limb solar flares and (2) the qualitative analysis of the model’s explanation indicates that our model identifies and uses features associated with ARs in central and near-limb locations from full-disk magnetograms to make corresponding predictions. In other words, our models learn the shape and texture-based characteristics of flaring ARs even when they are at near-limb areas, which is a novel and critical capability that has significant implications for operational forecasting. 

\keywords{Solar flares \and Deep learning \and Explainable AI.}
\end{abstract}
\section{Introduction}
Solar flares are temporary occurrences on the Sun that can generate abrupt and massive eruptions of electromagnetic radiation in its outermost atmosphere. These events happen when magnetic energy, accumulated in the solar atmosphere, is suddenly discharged, leading to a surge of energy that spans a wide range of wavelengths, from radio waves to X-rays. They are considered critical phenomena in space weather forecasting, and predicting solar flares is essential to understanding and preparing for their effects on Earth's infrastructure and technological systems. The National Oceanic and Atmospheric Administration (NOAA) classifies solar flares into five groups based on their peak X-ray flux level, namely A, B, C, M, and X, which represent the order of the flares from weakest to strongest \cite{spaceweather} and are commonly referred to as NOAA/GOES flare classes, where GOES stands for Geostationary Operational Environmental Satellite. M- and X-class flares, which are rare but significant, are the strongest flares that can potentially cause near-Earth impacts, including disruptions in electricity supply chains, airline traffic, satellite communications, and radiation hazards to astronauts in space. This makes them of particular interest to researchers studying space weather. Therefore, developing better methods to predict solar flares is necessary to prepare for the effects of space weather on Earth.

Active regions (ARs) are typically characterized by strong magnetic fields that are concentrated in sunspots. These magnetic fields can become highly distorted and unstable, leading to the formation of plasma instabilities and the release of energy in the form of flares and other events \cite{Toriumi2019}. Most operational flare forecasts target these regions of interest and issue predictions for individual ARs, which are the main initiators of space weather events. In order to produce a comprehensive forecast for the entire solar disk using an AR-based model, a heuristic function is used to combine the output flare probabilities ($P_{FL}(AR_i)$) for each active region (AR) \cite{Pandey2022f}. The resulting probability, $P_{aggregated}= 1 - \prod_{i}\big[1-{P_{FL}(AR_i)\big]}$, represents the likelihood of at least one AR producing a flare, assuming that the flaring events from different ARs are independent. However, there are two main issues with this approach for operational systems. Firstly, magnetic field measurements, which are the primary feature used by AR-based models, are subject to projection effects that distort measurements when ARs are closer to the limb. As a result, the aggregated full-disk flare probability is restricted to ARs in central locations, typically within $\pm$30$^{\circ}$ \cite{Huang2018}, $\pm$45$^{\circ}$ \cite{Li2020} to $\pm$70$^{\circ}$ of the disk center \cite{Ji2020}. Secondly, the heuristic function assumes that all ARs are equally important and independent of one another, which limits the accuracy of full-disk flare prediction probability. In contrast, full-disk models use complete magnetograms covering the entire solar disk, which are used to determine shape-based parameters such as size, directionality, borders, and inversion lines\cite{Ji2023}. Although projection effects still exist in these images, full-disk models can learn from the near-limb areas and provide a complementary element to AR-based models by predicting flares that occur in these regions \cite{Pandey2021}.

Machine learning and deep learning methods are currently being applied to predict solar flares, with experimental success and interdisciplinary collaboration from researchers in various fields \cite{Nishizuka_2017}, \cite{Nishizuka2018}, \cite{Huang2018}, \cite{Li2020}, \cite{Pandey2021}, \cite{Ji2022}, \cite{Whitman2022}. Although these approaches have improved image classification and computer vision, they learn complex data representations, resulting in so-called black-box models. The decision-making process of these models is obscured, which is crucial for operational forecasting communities. To address this issue, several attribution methods, or post hoc analysis methods, have been developed to explain and interpret the decisions made by deep neural networks. These methods focus on analyzing trained models and do not contribute to the model's parameters during training. In this study, we develop a convolutional neural network (CNN) based full-disk model for predicting solar flares with a magnitude of $\geq$M-class flares. We evaluate and explain the model's performance using three attribution methods: Guided Gradient-weighted Class Activation Mapping (Guided Grad-CAM) \cite{gradcam}, Integrated Gradients \cite{IntGrad}, and Deep Shapley Additive Explanations (Deep SHAP) \cite{DeepShap}. Our analysis reveals that our model's decisions are based on the characteristics corresponding to ARs, and it successfully predicts flares appearing on near-limb regions of the Sun.

The rest of this paper is organized as follows. In Sec.~\ref{sec:rel}, we present the related work on flare forecasting. In Sec.~\ref{sec:data}, we present our methodology with data preparation and model architecture. In Sec.~\ref{sec:attr}, we provide a detailed description of all three attribution methods used as methods of explanation. In Sec.~\ref{sec:eval}, we present our experimental evaluation. In Sec.~\ref{sec:dis}, we discuss the interpretation of our models, and in Sec.~\ref{sec:cf}, we present our conclusion and future work.

\section{Related Work}\label{sec:rel}
Currently, there are four main types of methods in use for predicting solar flares, which include (i) human-based prediction techniques based on empirical observations \cite{Crown2012}, \cite{Devos2014} (ii) statistical approaches \cite{Lee2012}, \cite{Leka2018} (iii) numerical simulations based on physics-based models \cite{Kusano2020}, \cite{Korss2020}, and (iv) data-driven models which made use of machine learning and deep learning techniques \cite{Bobra2015}, \cite{Huang2018}, \cite{Li2020}, \cite{Ahmadzadeh2019}, \cite{Pandey2021}, \cite{Pandey2022}. The application of machine learning in predicting solar flares has seen significant progress due to recent advances. In one such application of machine learning, a multi-layer perceptron model based on machine learning was employed for predicting $\geq$C- and $\geq$M-class flares in \cite{Nishizuka2018} using 79 manually selected physical precursors derived from multi-modal solar observations.

Later, a CNN-based model was developed for predicting $\geq$C-, $\geq$M-, and $\geq$X-class flares using solar AR patches extracted from line-of-sight (LoS) magnetograms within $\pm$30$^{\circ}$ of the central meridian in \cite{Huang2018}, taking advantage of the increasing popularity of deep learning models. \cite{Li2020} also used a CNN-based model to predict $\geq$C- and $\geq$M-class flares within 24 hours using AR patches located within $\pm45^{\circ}$ of the central meridian. To address the class imbalance issue, they employed undersampling and data augmentation techniques. However, while undersampling led to higher experimental accuracy scores, it often failed to deliver similar real-time performance \cite{Ahmadzadeh2021}. It is worth noting that both of these models have limited operational capability as they are restricted to a small portion of the observable disk in central locations ($\pm30^{\circ}$ and $\pm45^{\circ}$).

In addition, in \cite{Park2018}, a CNN-based hybrid model was introduced which combined GoogleLeNet \cite{Szegedy2015} and DenseNet \cite{Huang2017}. The model was trained using a large volume of data from both the Helioseismic and Magnetic Imager (HMI) instrument onboard Solar Dynamics Observatory (SDO) and magnetograms from the Michelson Doppler Imager (MDI) onboard the Solar and Heliospheric Observatory (SOHO). The aim of this model was to predict the occurrence of $\geq$C-class flares within the next 24 hours. However, it is important to note that these two instruments are not currently cross-calibrated for forecasting purposes, which may result in spurious or incomplete patterns being identified. More recently, an AlexNet-based \cite{alex} full-disk flare prediction model was presented in \cite{Pandey2021}. The authors provided a black-box model, but training and validation were limited due to a lower temporal resolution.

To interpret a CNN-based solar flare prediction model trained with AR patches, \cite{Bhattacharjee2020} used an occlusion-based method, and \cite{Yi2021} presented visual explanation methods using daily observations of solar full-disk LoS magnetograms at 00:00 UT. They applied Grad-CAM \cite{gradcam} and Guided Backpropagation \cite{gbackprop} to explore the relationship between physical parameters and the occurrence of C-, M-, and X-class flares. However, these methods had limitations in predicting near-limb flares. Recently, \cite{Sun2022} evaluated two additional attribution methods, DeepLIFT \cite{Deeplift} and Integrated Gradients \cite{IntGrad}, for interpreting CNNs trained on AR patches from central locations, i.e., within $\pm70^{\circ}$ for predicting solar flares.

In this paper, a CNN-based model is presented for predicting $\geq$M-class flares, which was trained using full-disk LoS magnetogram images. The contributions of this study are threefold: (i) demonstrating an overall improvement in the performance of a full-disk solar flare prediction model, (ii) utilizing recent attribution methods to provide explanations of our model's decisions, and (iii) for the first time, demonstrating the capability of predicting flares in near-limb regions of the Sun, which are traditionally difficult to predict with AR-based models.

\section{Data and Model}\label{sec:data}
\begin{figure}[ht]
\centering
\includegraphics[width=0.9\linewidth ]{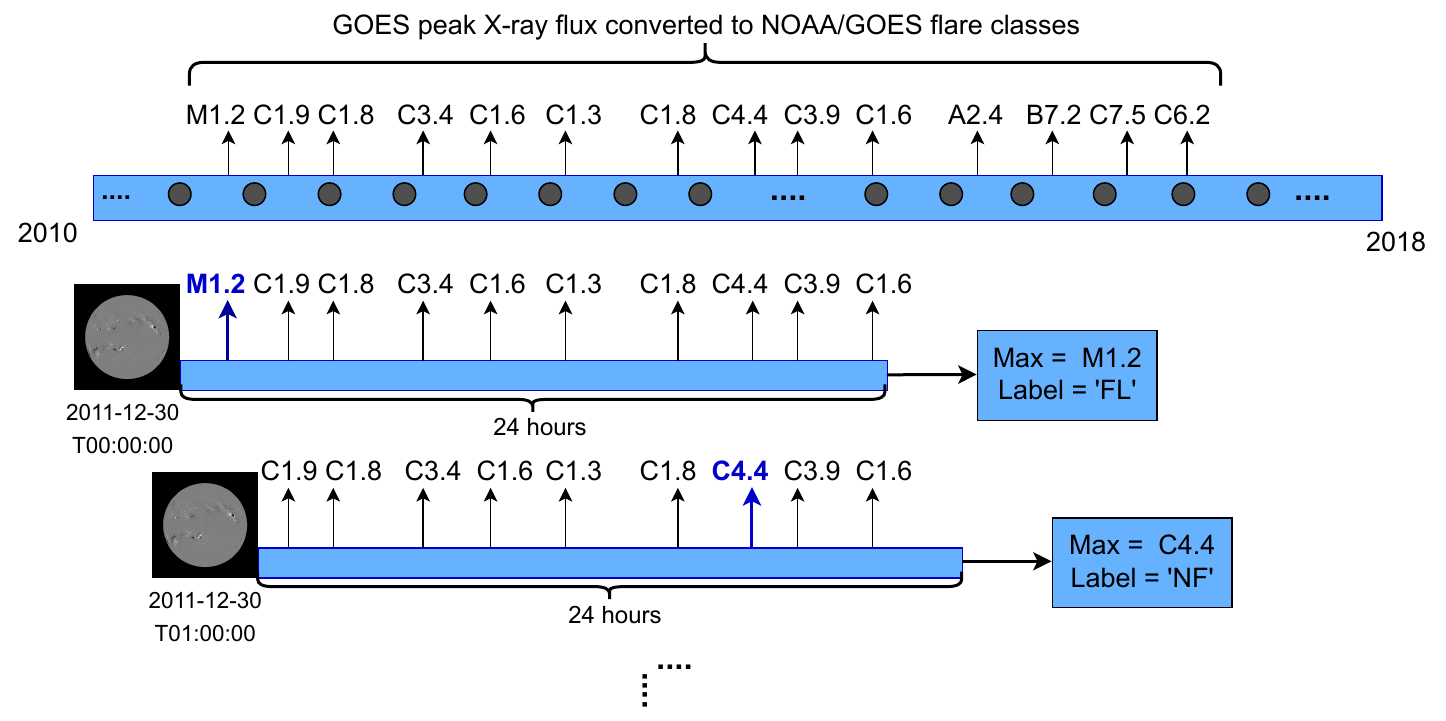}
\caption[]{A visual representation of the data labeling process using hourly observations of full-disk LoS magnetograms with a prediction window of 24 hours. Here, 'FL' and 'NF' indicates Flare and No Flare for binary prediction ($\geq$M-class flares). The gray-filled circles indicate hourly spaced timestamps for magnetogram instances.}
\label{fig:timeline}
\end{figure}
\begin{figure}[b!]
\centering
\includegraphics[width=0.98\linewidth ]{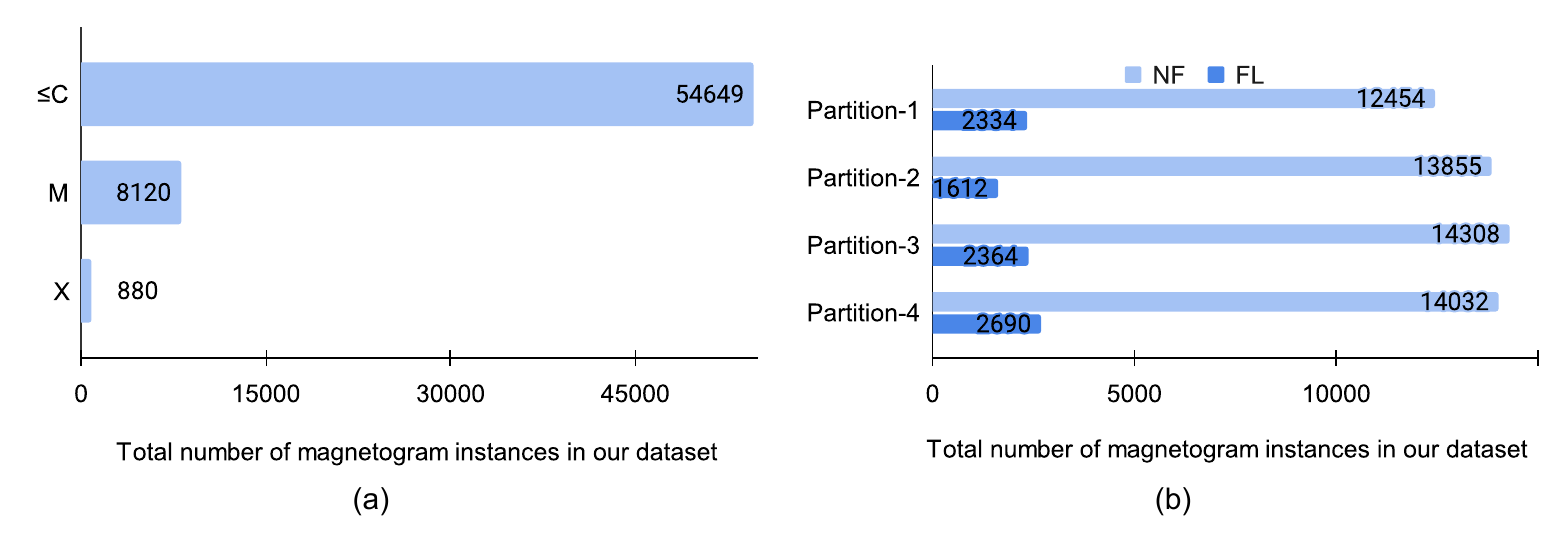}
\caption[]{(a) The total number of hourly sampled magnetograms images per flare classes.
(b) Label distribution into four tri-monthly partitions for predicting $\geq$M-class flares.}
\label{fig:partitions}
\end{figure}

We used compressed images of full-disk LoS solar magnetograms obtained from the HMI/SDO available in near real-time publicly via Helioviewer\footnote[1]{\url{Helioviewer API V2: https://api.helioviewer.org/docs/v2/}} \cite{Muller2009}. We sampled the magnetogram images every hour of the day, starting at 00:00 and ending at 23:00, from December 2010 to December 2018. We collected a total of 63,649 magnetogram images and labeled them using a 24-hour prediction window based on the maximum peak X-ray flux (converted to NOAA/GOES flare classes) within the next 24 hours, as illustrated in Fig.~\ref{fig:timeline}. To elaborate, if the maximum X-ray intensity of a flare was weaker than M (i.e., $<10^{-5}Wm^{-2}$), we labeled the observation as "No Flare" (NF: $<$M), and if it was $\geq$M, we labeled it as "Flare" (FL: $\geq$M). This resulted in 54,649 instances for the NF class and 9,000 instances for the FL class. The detailed class-wise distribution of our data is shown in Fig.~\ref{fig:partitions}(a). Finally, we created a non-chronological split of our data into four temporally non-overlapping tri-monthly partitions for our cross-validation experiments. We created this partitioning by dividing the data timeline from December 2010 to December 2018 into four partitions. Partition-1 contained data from January to March, Partition-2 contained data from April to June, Partition-3 contained data from July to September, and Partition-4 contained data from October to December, as shown in Fig.~\ref{fig:partitions}(b). Due to the scarcity of $\geq$M-class flares, the overall distribution of the data is highly imbalanced, with FL:NF $\sim$1:6.

\begin{figure}[tbh!]
  \centering
  \includegraphics[width=0.9\linewidth]{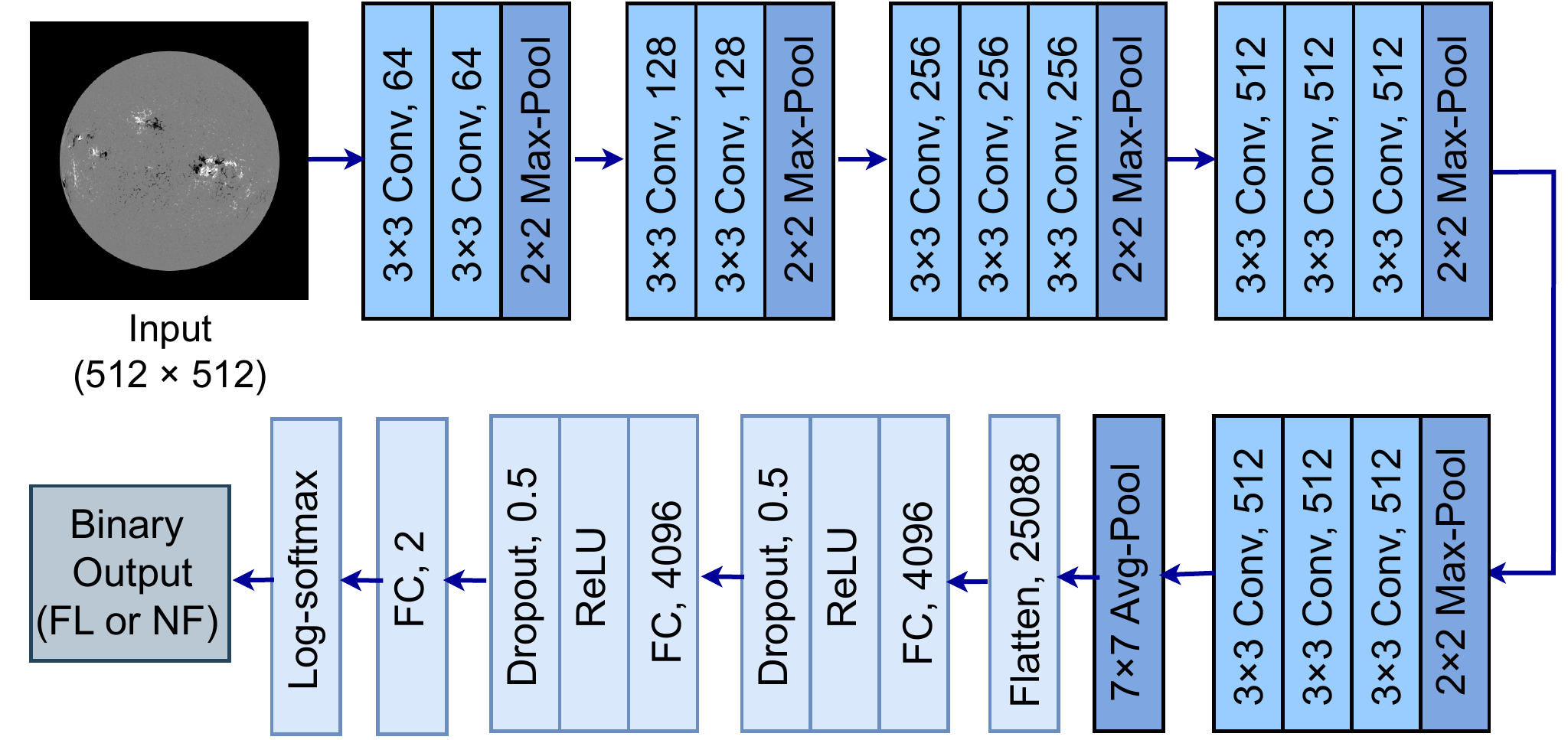}
  \caption{The architecture of our full-disk solar flare prediction model.}
  \label{fig:arch1}
\end{figure}

In our study, we employed transfer learning with a pre-trained VGG-16 model \cite{vgg16} for solar flare prediction. To use the pre-trained weights for our 1-channel input magnetogram images, we duplicated the channels twice, as the pre-trained model requires a 3-channel image for input. Additionally, we used the 7$\times$7 adaptive average pooling after feature extraction using the convolutional layer and prior to the fully-connected layer to match the dimension of our 1-channel, 512$\times$512 image. This ensures efficient utilization of the pre-trained weights, irrespective of the architecture of the VGG-16 model, which is designed to receive 224$\times$224, 3-channel images. Our model comprises 13 convolutional layers, each followed by a rectified linear unit (ReLU) activation, five max pool layers, one average pool layer, and three fully connected layers, as illustrated in Fig.~\ref{fig:arch1}.

\section{Attribution Methods}\label{sec:attr}

Deep learning models are often seen as black boxes due to their intricate data representations, making them difficult to understand, leading to issues of inconsistency in the discovered patterns \cite{e23010018}. The attribution methods are post hoc approaches for model interpretation that provides insights into the decision-making process of the trained CNN models without influencing the training process. These methods generate an attribution vector, or heat map, of the same size as the input, where each element in the vector represents the contribution of the corresponding input element to the model's decision. Attribution methods can be broadly classified into two main categories: perturbation-based and gradient-based \cite{Holzinger2022}. Perturbation-based methods modify the parts of the input to create new inputs and compute the attribution by measuring the difference between the output of the original and modified inputs. However, this approach can lead to inconsistent interpretations due to the creation of Out-of-Distribution (OoD) data caused by random perturbations \cite{Qiu2022}. In contrast, gradient-based methods calculate the gradients of the output with respect to the extracted features or input using backpropagation, enabling attribution scores to be estimated more efficiently and robustly to input perturbations \cite{Nielsen2022}. 

Therefore, in this study, we employed three recent gradient-based methods to evaluate our models due to their reliability and computational efficiency. Our primary objective is to provide a visual analysis of the decisions made by our model and identify the characteristics of magnetogram images that trigger specific decisions by cross-validating the generated explanations from all three methods, which can clarify the predictive output of the models and help with operational forecasting under critical conditions.

\noindent\textbf{Guided Grad-CAM: }
The Guided Gradient-weighted Class Activation Mapping (Guided Grad-CAM) method \cite{gradcam} combines the strengths of Grad-CAM and guided backpropagation \cite{gbackprop}. Grad-CAM produces a coarse localization map of important regions in the image by using class-specific gradient information from the final convolutional layer of a CNN, while guided backpropagation calculates the gradient of the output with respect to the input, highlighting important pixels detected by neurons. While Grad-CAM attributions are class-discriminative and useful for localizing relevant image regions, they do not provide fine-grained pixel importance like guided backpropagation \cite{Chattopadhay2018}. Guided Grad-CAM combines the fine-grained pixel details from guided backpropagation with the coarse localization advantages of Grad-CAM and generates its final localization map by performing an element-wise multiplication between the upsampled Grad-CAM attributions and the guided backpropagation output.

\noindent\textbf{Integrated Gradients: }
Integrated Gradients (IG) \cite{IntGrad} is an attribution method that explains a model's output by analyzing its features. To be more specific, IG calculates the path integral of gradients along a straight line connecting the baseline feature to the input feature in question. A baseline reference is required for this method, which represents the absence of a feature in the original image and can be a zero vector or noise; we used a zero vector of the size of the input as a baseline for our computation. IG is preferred for its completeness property, which states that the sum of integrated gradients for all features equals the difference between the model's output with the given input and the baseline input values. This property allows for attributions to be assigned to each individual feature and, when added together, should yield the output value itself \cite{Sturmfels2020}.

\noindent\textbf{Deep SHAP: }
SHAP values, short for SHapley Additive exPlanations \cite{DeepShap}, utilize cooperative game theory \cite{1952} to enhance the transparency and interpretability of machine learning models. This method quantifies the contribution or importance of each feature on the model's prediction rather than evaluating the quality of the prediction itself. In the case of deep-learning models, Deep SHAP \cite{DeepShap} improves upon the DeepLIFT algorithm \cite{Deeplift} by estimating the conditional expectations of SHAP values using a set of background samples. For each input sample, the DeepLIFT attribution is computed with respect to each baseline, and the resulting attributions are averaged. This method assumes feature independence and explains the model's output through the additive composition of feature effects. Although it assumes a linear model for each explanation, the overall model across multiple explanations can be complex and non-linear. Similar to IG, Deep SHAP also satisfies the completeness property \cite{Sturmfels2020}.

\section{Experimental Evaluation}\label{sec:eval}
\subsection{Experimental Settings}
We trained a full-disk flare prediction model using Stochastic Gradient Descent (SGD) as an optimizer and Negative Log-Likelihood (NLL) as the objective function. To apply NLL loss, we used logarithmic-softmax activation on the raw logits from the output layer. Our model was initialized with pre-trained weights from the VGG-16 model \cite{vgg16}. We further trained the model for 50 epochs with a batch size of 64 using dynamic learning rates (initialized at 0.001 and halved every 5 epochs). To address the class imbalance issue, we used data augmentation and class weights in the loss function. Specifically, we applied three augmentation techniques (vertical flipping, horizontal flipping, and rotations of +5$^{\circ}$ to -5$^{\circ}$) during the training phase to explicitly augment the minority FL-class three times. However, this still left the dataset imbalanced, so we adjusted the class weights inversely proportional to the class frequencies after augmentations and penalized misclassifications made in the minority class. To improve the generalization of our model without introducing bias in the test set, we applied data augmentation exclusively during the training phase, and we opted for augmentation over oversampling and undersampling as the latter two may lead to overfitting of the model \cite{Ahmadzadeh2019}. Finally, we conducted 4-fold cross-validation experiments using tri-monthly partitions to train our models. 

We assess the overall performance of our models using two forecast skills scores: True Skill Statistics (TSS, in Eq.~\ref{eq:TSS}) and Heidke Skill Score (HSS, in Eq.~\ref{eq:HSS}), derived from the elements of confusion matrix: True Positive (TP), True Negative (TN), False Positive (FP), and False Negative (FN). In this context, FL and NF represent positive and negative classes respectively. 

\begin{equation}\label{eq:TSS}
    TSS = \frac{TP}{TP+FN} - \frac{FP}{FP+TN} 
\end{equation}

\begin{equation}\label{eq:HSS}
    HSS = 2\times \frac{TP \times TN - FN \times FP}{((P \times (FN + TN) + (TP + FP) \times N))}
\end{equation}

where N = TN + FP and  P = TP + FN. TSS and HSS values range from -1 to 1, where 1 indicates all correct predictions, -1 represents all incorrect predictions, and 0 represents no skill. In contrast to TSS, HSS is an imbalance-aware metric, and it is common practice to use HSS for the solar flare prediction models due to the high class-imbalance ratio present in the datasets and for a balanced dataset, these metrics are equivalent as discussed in \cite{Ahmadzadeh2021}. Lastly, we report the subclass and overall recall for flaring instances (M- and X-class), which is calculated as ($\frac{TP}{TP+FN}$), to demonstrate the prediction sensitivity.  To reproduce this work, the source code and detailed experimental results can be accessed from our open source repository \footnote[2]{\url{explainFDvgg16: https://bitbucket.org/gsudmlab/explainfdvgg16/src/main/}}.

\subsection{Evaluation}
We performed 4-fold cross-validation using the tri-monthly dataset for evaluating our models. Our models have on average TSS$\sim$0.51 and HSS$\sim$0.35, which improves over the performance of \cite{Pandey2021} by $\sim$4\% in terms of TSS (reported $\sim$0.47) and competing results in terms of HSS (reported $\sim$0.35). In addition, we evaluate our results for correctly predicted and missed flare counts for class-specific flares (X-class and M-class) in central locations (within $\pm$70$^{\circ}$) and near-limb locations (beyond $\pm$70$^{\circ}$) of the Sun as shown in Table \ref{table:comp}. We observe that our models made correct predictions for $\sim$89\% of the X-class flares and $\sim$77\% of the M-class flares in central locations. Similarly, our models show a compelling performance for flares appearing on near-limb locations of the Sun, where $\sim$77\% of the X-class and $\sim$52\%  of the M-class flares are predicted correctly. This is important because, to our knowledge, the prediction of near-limb flares is often overlooked. More false positives in M-class are expected because of the model's inability to distinguish bordering class [C4+ to C9.9] flares from $\geq$M-class flares, which we have observed empirically in our prior work \cite{Pandey2022} as well. Overall, we observed that $\sim$86\% and $\sim$70\% of the X-class and M-class flares, respectively, are predicted correctly by our models.

\begin{table}[b!]
\setlength{\tabcolsep}{4pt}
\renewcommand{\arraystretch}{1.5}
\caption{Counts of correctly (TP) and incorrectly (FN) classified X- and M-class flares in central ($|longitude|$$\leq\pm70^{\circ}$) and near-limb locations. The recall across different location groups is also presented. Counts are aggregated across folds.}
\begin{center}
 \begin{tabular}{r c c c c c c}
\hline
 & 
\multicolumn{3}{c}{Within $\pm$70$^{\circ}$} 
&                                            
\multicolumn{3}{c}{Beyond $\pm$70$^{\circ}$}\\
Flare-Class & TP  & FN  & Recall  & TP   &FN & Recall \\
\hline
X-Class  &  597  & 71  & 0.89 & 164 & 48 & 0.77\\

M-Class &  4,464 & 1,366  & 0.77 & 1,197   &1,093 & 0.52\\

Total (X\&M) & 5,061 & 1,437 &0.78& 1,361 & 1,141 & 0.54\\ 
\hline
\end{tabular}
\end{center}
\label{table:comp}
\end{table}

\begin{figure}[ht!]
\centering
\includegraphics[width=0.95\linewidth ]{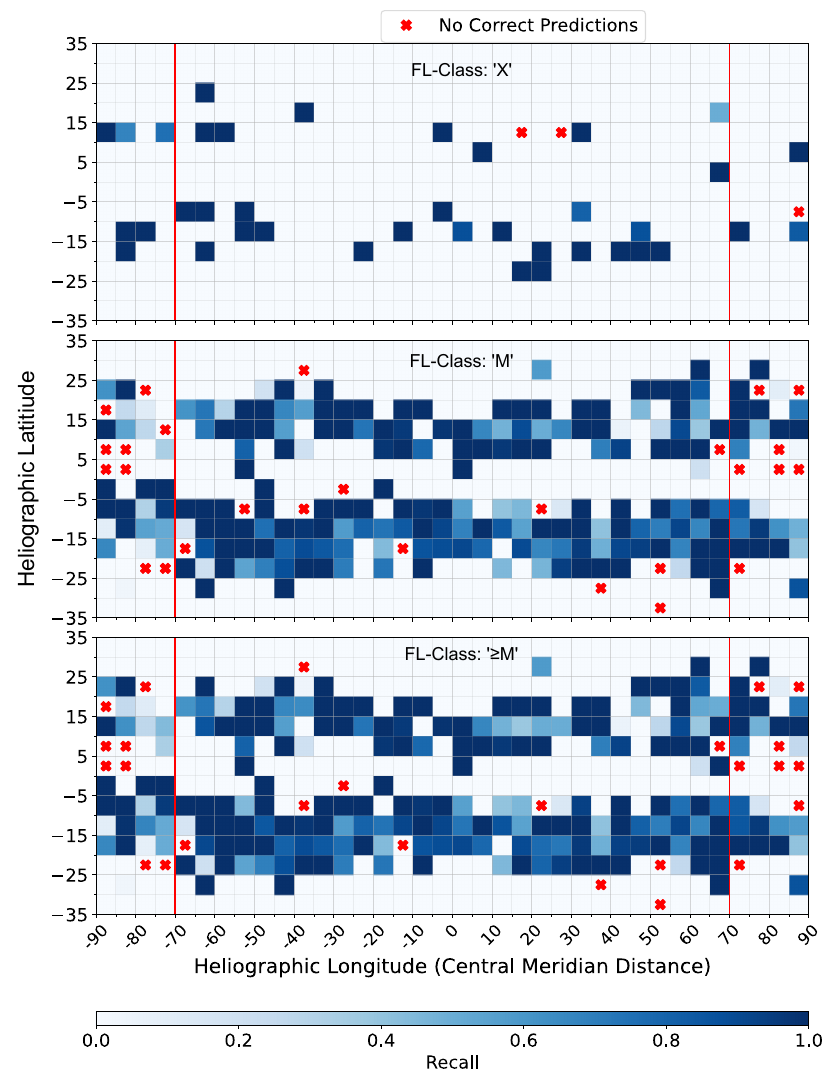}
\caption[]{A heat map showcasing recall for individual FL-Class (X- and M-class flares) and when combined ($\geq$M-class flares) binned into 5$^{\circ}$ $\times$ 5$^{\circ}$ flare locations used as the label. The flare events beyond $\pm$70$^{\circ}$ longitude (separated by a vertical red line) represent near-limb events.  Note: Red cross in white grids represents locations with zero correct predictions while white cells without red cross represent unavailable instances.}
\label{fig:histogram}
\end{figure}
We also quantitatively and qualitatively evaluated our models' effectiveness by spatially analyzing their performance with respect to the locations of M- and X-class flares responsible for the labels. To conduct our analysis, we have spatially binned the responsible flares (maximum X-ray flux within the next 24h) and analyzed whether these instances were correctly (TP) or incorrectly predicted (FN). For this, we used the predictions of our models in the validation set from the 4-fold cross-validation experiments. Here, each bin represents a 5$^{\circ}$ by 5$^{\circ}$ spatial cell in Heliographic Stonyhurst (HGS) coordinate system (i.e., latitude and longitude). For each subgroup, represented in a spatial cell, we calculate the recall for M-class, X-class, and M- and X-class flares, separately to assess the models' sensitivity at a fine-grained level. The heatmaps demonstrating the spatial distribution of recall scores of our models can be seen in Fig.~\ref{fig:histogram}. This allows us to pinpoint the locations where our models were more effective in making accurate predictions and vice versa. We observed that our models demonstrated reasonable performance overall, particularly for X-class flares, in both near-limb and central locations. However, we also observed a higher number of false negatives around near-limb locations for M-class flares. In particular, we demonstrate that the full-disk model proposed in this paper can predict flares appearing at near-limb locations of the Sun with great accuracy, which is a crucial addition to operational forecasting systems.
\begin{figure}[ht!]
  \centering
  \includegraphics[width=0.95\linewidth]{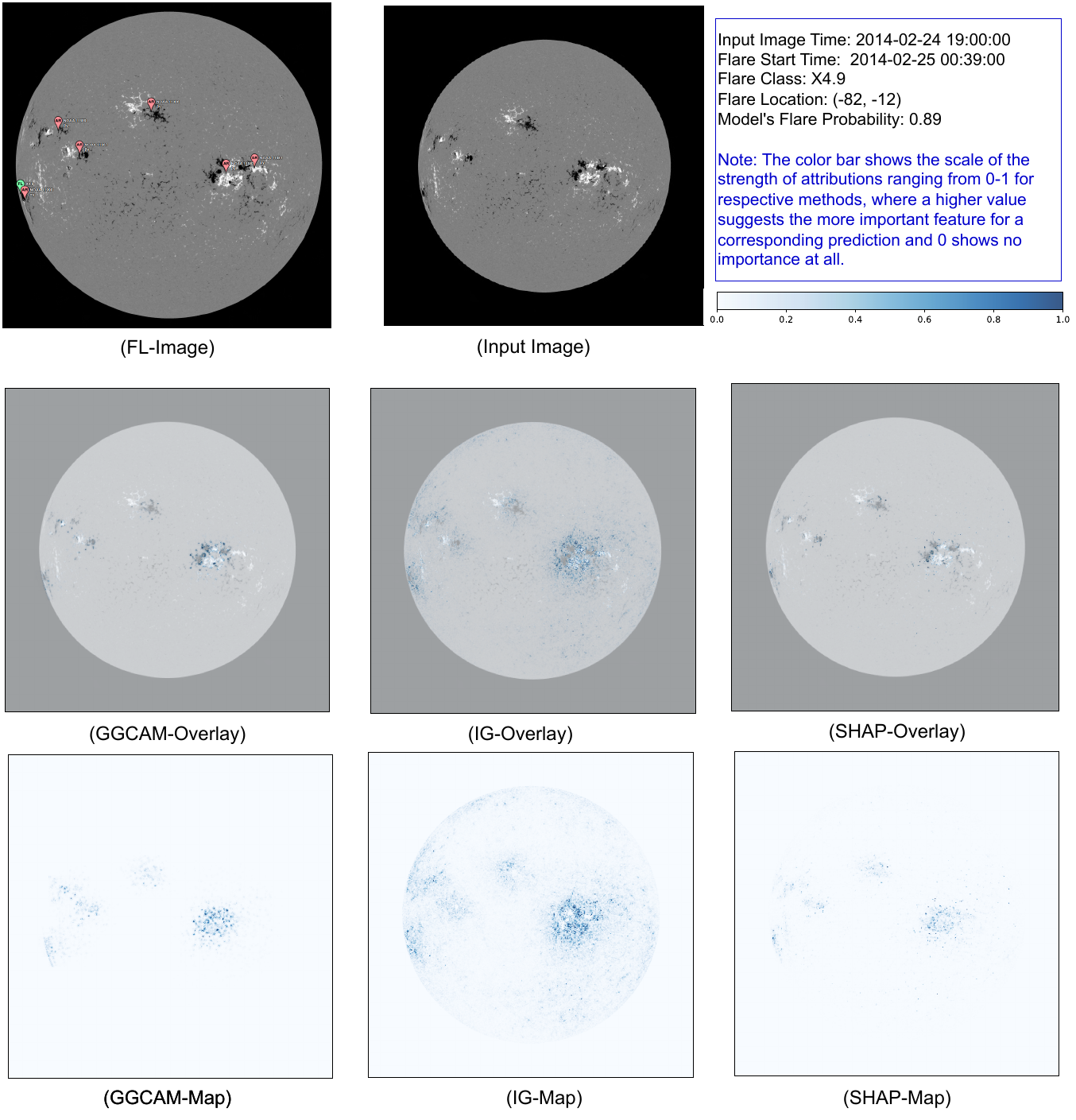}
  \caption{A visual explanation of correctly predicted near-limb (East) FL-class instance. (FL-Image): Annotated full-disk magnetogram at flare start time, showing flare location (green flag) and NOAA ARs (red flags). (Input Image):  Actual magnetogram from the dataset. Overlays (GGCAM, IG, SHAP) depict the input image overlayed with attributions, and Maps (GGCAM, IG, SHAP) showcase the attribution maps obtained from Guided Grad-CAM, Integrated Gradients, and Deep SHAP, respectively.}
  \label{fig:east}
\end{figure}

\section{Discussion}\label{sec:dis}

\begin{figure}[ht!]
  \centering
  \includegraphics[width=0.95\linewidth]{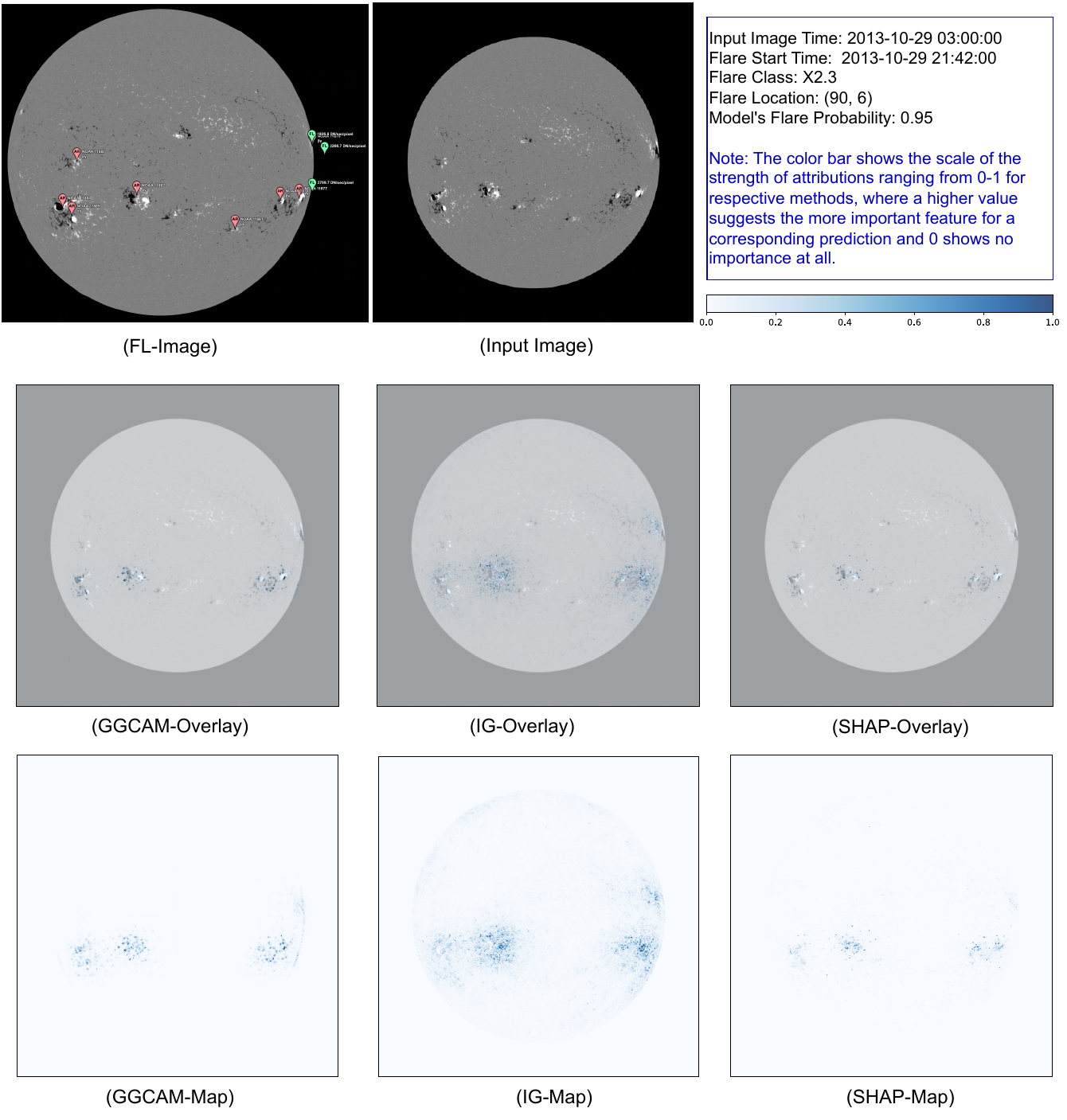}
  \caption{A visual explanation of correctly predicted near-limb (West) FL-class instance. (FL-Image): Annotated full-disk magnetogram at flare start time, showing flare location (green flag) and NOAA ARs (red flags). (Input Image):  Actual magnetogram from the dataset. Overlays (GGCAM, IG, SHAP) depict the input image overlayed with attributions, and Maps (GGCAM, IG, SHAP) showcase the attribution maps obtained from Guided Grad-CAM, Integrated Gradients, and Deep SHAP, respectively.}
  \label{fig:west}
\end{figure}
\begin{figure}[ht!]
  \centering
  \includegraphics[width=0.95\linewidth]{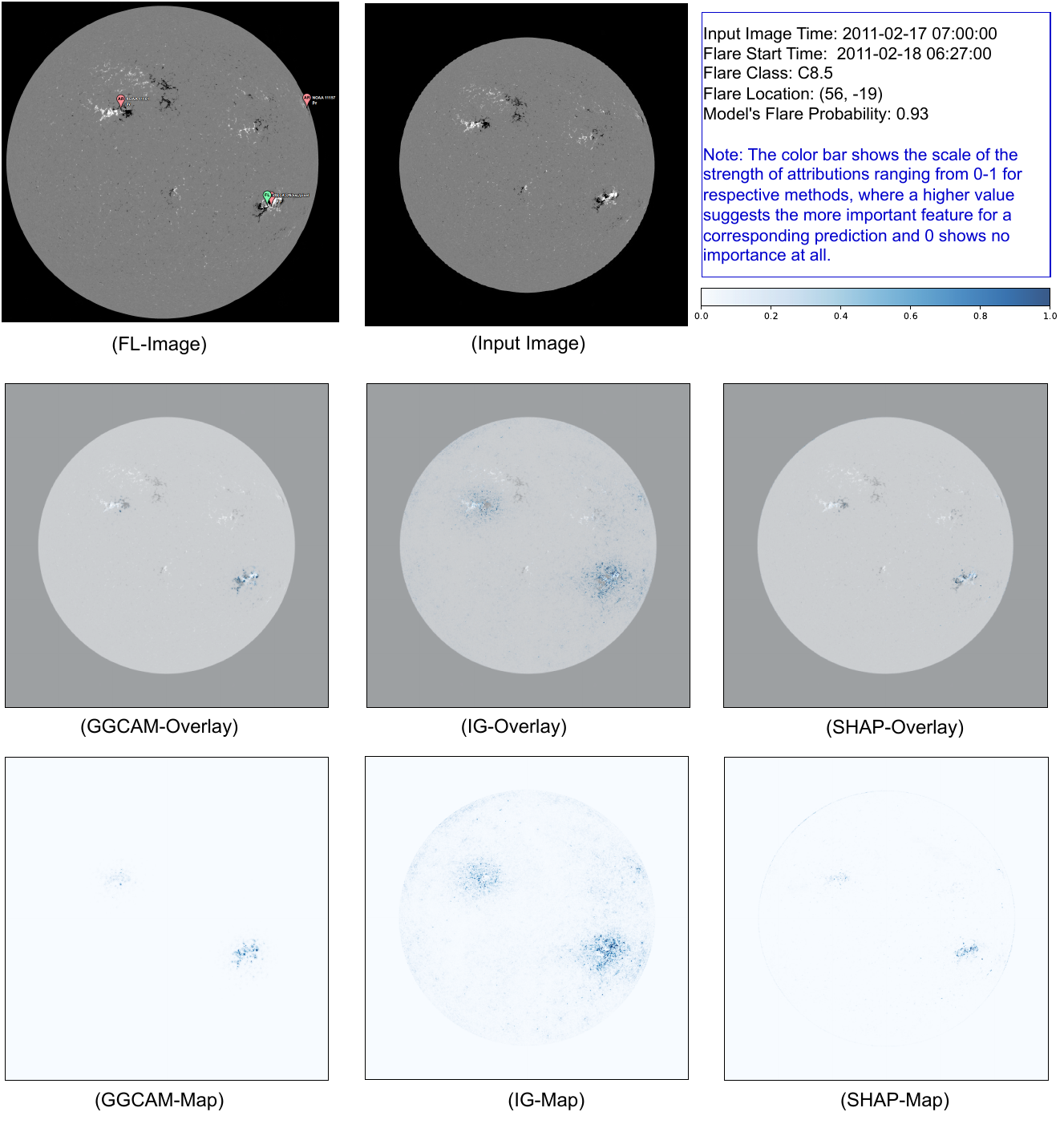}
  \caption{A visual explanation of incorrectly predicted NF-class instance. (FL-Image): Annotated full-disk magnetogram at flare start time, showing flare location (green flag) and NOAA ARs (red flags). (Input Image):  Actual magnetogram from the dataset. Overlays(GGCAM, IG, SHAP) depict the input image overlayed with attributions, and Maps(GGCAM, IG, SHAP)) showcase the attribution maps obtained from Guided Grad-CAM, Integrated Gradients, and Deep SHAP, respectively.}
  \label{fig:cflare}
\end{figure}
In this section, we interpret the visual explanations generated using the attribution methods mentioned earlier for correctly predicted near-limb flares and the model's high confidence in an incorrect prediction. As the major focus of this study is on the near-limb flares, we interpret the predictions of our model for an east-limb X4.9-class (note that East and West are reversed in solar coordinates) flare observed on 2014-02-25 at 00:39:00 UTC with a visual explanation generated using all three attribution methods. For this, we used an input image at 2014-02-24 19:00:00 UTC ($\sim$6 hours prior to the flare event), where the sunspot for the corresponding flare becomes visible in the magnetogram image. We observed while all three methods highlight features corresponding to an AR in the magnetogram, Guided Grad-CAM and Deep SHAP provide finer details by suppressing noise compared to IG as shown in Fig.~\ref{fig:east}. Furthermore, the visualization of attribution maps suggests that for this particular prediction, although barely visible, the region responsible for the flare event is considered important and hence contributes to the consequent decision. The explanation shows that as soon as a region becomes visible, the pixels covering the AR on the east-limb are activated. Similarly, we analyze another case of correctly predicted near-limb flare (West-limb) of the Sun. For this, we provide a case of X2.3-class flare observed on 2013-10-29T21:42:00 UTC where we used an input image at 2013-10-29T03:00:00 UTC ($\sim$19 hours prior to the flare event) shown in Fig.~\ref{fig:west}. We observed that the model focuses on specific ARs including the relatively smaller AR on the west limb, even though other ARs are present in the magnetogram image. This shows that our models are capable of identifying the relevant AR even when there is a severe projection effect.

Similarly, to analyze a case of false positive, we present an example of a C8.5 flare observed on 2011-02-18 at 06:27:00 UTC, and to explain the result, we used an input magnetogram instance at 2014-02-17 07:00:00 UTC ($\sim$23.5 hours prior to the event). We observed that the model's flaring probability for this particular instance is about 0.93. Therefore, we seek a visual explanation of this prediction using all three interpretation methods. Similar to the observations from our positive prediction, the visualization rendered using Guided Grad-CAM and Deep SHAP provides smoothed details and reveals that out of three ARs present in the magnetogram, only two of them are activated while the AR on the west-limb is not considered important for this prediction as shown in Fig.~\ref{fig:cflare}. Although an incorrect prediction, the visual explanation shows that the model's decision is based on an AR which is, in fact, responsible for the eventual C8.5 flare event. This incorrect prediction can be attributed to the interference of these bordering class flares, which is problematic for binary flare prediction models. 
\section{Conclusion and Future Work}\label{sec:cf}
In this paper, we employed three recent gradient-based attribution methods to interpret the predictions made by our binary flare prediction model based on VGG-16, which was trained to predict $\geq$M-class flares. We addressed the issue of flares occurring in near-limb regions of the Sun, which has been widely ignored, and our model demonstrated competent performance for such events. Additionally, we assessed the model's predictions with visual explanations, indicating that the decisions were primarily based on characteristics related to ARs in the magnetogram instance. Despite the model's enhanced ability, it still suffers from a high false positive rate due to high C-class flares. In an effort to address this problem, we plan to examine the unique features of each flare class to create a more effective method for segregating these classes based on background flux and generate a new set of labels that better handle border class flares. Moreover, our models currently only examine spatial patterns in our data, but we intend to broaden this work to include spatiotemporal models to improve performance.\\

\section*{Acknowledgements}
This work is supported in part under two NSF awards \#2104004 and \#1931555, jointly by the Office of Advanced Cyberinfrastructure within the Directorate for Computer and Information Science and Engineering, the Division of Astronomical Sciences within the Directorate for Mathematical and Physical Sciences, and the Solar Terrestrial Physics Program and the Division of Integrative and Collaborative Education and Research within the Directorate for Geosciences. This work is also partially supported by the National Aeronautics and Space Administration (NASA) grant award \#80NSSC22K0272.

\section*{Ethical Statement}
Space weather forecasting research raises several ethical implications that must be considered. It is important to note that the data used for the full-disk deep learning model for solar flare prediction is publicly available as a courtesy of NASA/SDO and the AIA, EVE, and HMI science teams -- and not subject to data privacy and security concerns. The use of SDO images for non-commercial purposes and public education and information efforts is strongly encouraged and requires no expressed authorization. However, it is still essential to consider the ethical implications associated with developing and using a full-disk deep learning model for solar flare prediction, particularly in terms of fairness, interpretability, and transparency. It is crucial to ensure that the model is developed and used ethically and responsibly to avoid any potential biases or negative impacts on individuals or communities. Moreover, post hoc analysis for full-disk deep learning models for solar flare prediction should avoid giving wrongful assumptions of causality and false trust. While these models may have robust and novel forecast skills, it is crucial to understand the scarcity of extreme solar events and the skill scores used to assess model performance. We note that these models are not perfect and have limitations that should be considered when interpreting their predictions. Therefore, it is important to use these models with caution and to consider multiple sources of information when making decisions, especially when in operations, related to space weather events. By being transparent about the limitations and uncertainties associated with these models, we can ensure that they are used ethically and responsibly to mitigate any potential harm to individuals or communities.

Furthermore, the impact of space weather events can range from minor disruptions to significant damage to critical infrastructure, such as power grids, communication systems, and navigation systems, with the potential to cause significant economic losses. Therefore, it is crucial to ensure public safety, particularly for astronauts and airline crew members, by providing information about potential dangers associated with space weather events. Finally, it is imperative to ensure that space weather forecasting research is used for peaceful purposes, i.e., early detection and in part avoiding vulnerabilities that may be caused by extreme space weather events.
\bibliographystyle{splncs04}
\bibliography{references}
\end{document}